\documentclass{article}

\usepackage{arxiv}

\usepackage[utf8]{inputenc} % allow utf-8 input
\usepackage[T1]{fontenc}    % use 8-bit T1 fonts
\usepackage{hyperref}       % hyperlinks
\usepackage{url}            % simple URL typesetting
\usepackage{booktabs}       % professional-quality tables
\usepackage{amsfonts}       % blackboard math symbols
\usepackage{nicefrac}       % compact symbols for 1/2, etc.
\usepackage{microtype}      % microtypography
\usepackage{lipsum}		% Can be removed after putting your text content
\usepackage{graphicx}
\usepackage{natbib}
\usepackage{doi}

\usepackage{mathtools}
\usepackage{multirow}
\usepackage{xcolor}
\usepackage{amssymb}
\usepackage{adjustbox}
\usepackage{subfig}
\usepackage{siunitx}
\usepackage{booktabs}

\newcommand{\norm}[1]{\left\lVert#1\right\rVert}

\title{On the Effectiveness of Generative Adversarial Network on Anomaly Detection}

\author{\hspace{-20pt}Laya Rafiee Sevyeri \qquad \hspace{40pt}Thomas Fevens \\
  \texttt{laya.rafiee@gmail.com} \qquad \texttt{fevens@cs.concordia.ca}\\
  Gina Cody School of Engineering and Computer Science\\
  Concordia University\\
  Montr\'eal, QC, Canada \\}

\date{}

\renewcommand{\shorttitle}{On the Effectiveness of Generative Adversarial Network on Anomaly Detection}

\hypersetup{
pdftitle={A template for the arxiv style},
pdfsubject={q-bio.NC, q-bio.QM},
pdfauthor={David S.~Hippocampus, Elias D.~Striatum},
pdfkeywords={First keyword, Second keyword, More},
}

\begin{document}
\maketitle

\begin{abstract}
	Identifying anomalies refers to detecting samples that do not resemble the training data distribution. Many generative models have been used to find anomalies, and among them, generative adversarial network (GAN)-based approaches are currently very popular. GANs mainly rely on the rich contextual information of these models to learn the actual training distribution. Following this analogy, we suggested a new unsupervised model based on GANs --a combination of an autoencoder and a GAN. Further, a new scoring function was introduced to target anomalies where a linear combination of the internal representation of the discriminator and the generator's visual representation, plus the encoded representation of the autoencoder, come together to define the proposed anomaly score. The model was further evaluated on benchmark datasets such as SVHN, CIFAR10, and MNIST, as well as a public medical dataset of leukemia images. In all the experiments, our model outperformed its existing counterparts while slightly improving the inference time.
\end{abstract}

% keywords can be removed

\section{Introduction}
Anomaly detection (AD), or sometimes novelty detection, outlier detection, or in a broad description out-of-distribution detection, is an interesting and well-known research topic that is widely studied in many fields such as network intrusion \citep{leung2005unsupervised}, fraud detection \citep{fawcett1997adaptive}, and computer vision \citep{mahadevan2010anomaly}. The problem focuses on identifying samples that deviate from other observations on data, indicating variability in measurement, experimental errors, or a novelty. In other words, finding those samples that do not fit the training data distribution is known as anomaly detection. This can be helpful to identify unknown anomalies in the medical domain where finding an appropriate annotated dataset is always a concern. This approach is also applicable in cases where the knowledge regarding the type of anomalies is limited.

A generative adversarial network (GAN)~\citep{goodfellow2014generative} has two components, a generator and a discriminator, with a multi-objective optimization which forms a zero-sum game between these two components leading to rich representations of the training data where these representations can be further utilized for downstream tasks. Generating realistic images of natural images~\citep{Radford2016dcgan, karras2019styleGAN, karras2020analyzing, karras2021alias} and medical images~\citep{han2018ganMRI}, image-to-image translation~\citep{isola2017imageToimageTranslation, zhu2017cycleGAN}, and text-to-image translation~\citep{zhang2017stackgan, dash2017tac, reed2016learnTodraw, lao2019dual} are some of the recent practices that achieved state-of-the-art performance using the idea of GAN. Aside from the fact that GANs can model the training distribution, using them to identify anomalies needs finding the corresponding latent representation of a given test image which is not obtained easily. Previous studies suggested either optimizing the input noise to the GANs~\citep{schlegl2017unsupervised} or using another module trained alongside the GAN~\citep{zenati2018efficient, zenati2018adversarially} to obtain the desired representation.

Following the importance of detecting anomalies in both natural and medical images, we present a simple and effective model based on GANs. In this model, a GAN and an autoencoder train simultaneously to learn the desired representations of the normal samples, which further will be used to indicate anomalies. In this work, anomalies are detected based on a new scoring function--a modification on previous anomaly score by considering multiple representations of a single image obtained from a GAN and an autoencoder. The experimental results on various domains; natural images (MNIST, CIFAR10, and SVHN), and medical imaging (Acute Lymphoblastic Leukemia (ALL)~\citep{labati2011all}) datasets demonstrate that our suggested generative model is capable of identifying anomalous (out-of-distribution) samples in different settings. Our model not only improved all the existing models in all the experiments but also showed that even if it trained on a very small dataset, the representations are rich enough to target anomalies. All the conducted experiments and the code are available \href{https://github.com/LayaaRS/Unsupervised-Anomaly-Detection-with-a-GAN-Augmented-Autoencoder}{\color{blue} here}.

% \url{https://github.com/LayaaRS/Unsupervised-Anomaly-Detection-with-a-GAN-Augmented-Autoencoder}.

\section{Related Work}
There are numerous different approaches in the literature to identify anomalies in various domains. In the context of images, these studies can be divided into three sub-categories. 1) The first category of research considers classical machine learning (ML) approaches such as one-class support vector machines (SVMs)~\citep{tax2004support} and clustering~\citep{xiong2011group} to detect anomalies. 2) The second type of work, also known as hybrid models, combine the classical ML and deep learning models; e.g., a one-class SVM on top of deep belief networks (DBNs)~\citep{erfani2016high} or an autoencoder with a k-means clustering on top~\citep{aytekin2018clustering}. 3) The last category includes recent develops in deep learning and designed purely based on the representations they provide. Variational autoencoders~\citep{an2015variational} and autoencoders~\citep{zhou2017anomaly} showcase the power of deep models for detecting anomalies.

In the last category, there is a series of work that has been leveraging GANs to obtain the desired representations for the purpose of detecting anomalies. However, finding meaningful representations of the distribution of the normal images is a challenging task. In one of the very first works,~\citet{schlegl2017unsupervised} proposed AnoGAN, a vanilla GAN accompanied with an optimization process on latent representation during inference procedure, to detect anomalies in the medical domain. A year later,~\citet{zenati2018efficient, zenati2018adversarially} proposed two different models based on BiGAN~\citep{Donahue2017bigan}, the recently proposed feature learning model, for the task of anomaly detection with a significant improvement on the inference time.

Following the recent successes using GANs and their variations on AD tasks, we introduce an unsupervised model based on GAN. Our model contains two generative models, a GAN and an autoencoder, to obtain the desired representation of a given image with two purposes, improving the performance of existing unsupervised AD models, and decreasing the detection time. A preliminary version of this study appears in~\citep{rafiee2020unsupervised}.

\section{Anomaly Detection}\label{sec:anomaly_detection}
The idea of using a GAN to find anomalies can be divided into two steps; learning the corresponding latent representation of a given image and a distance metric on how far the generated output is from the given image.

Previous studies each took advantage of GANs differently with a tailored distance metric for their proposed model to identify anomalies. In the next section, the similarities and differences of each of these two steps in the previous GAN-based models will be briefly described. Later on, the details of our AD model will be explained.

\subsection{GANs for Anomaly Detection}\label{sec:anogan}
Following the success of GANs and their application in various domains,~\citet{schlegl2017unsupervised} introduced the first anomaly detection model based on GANs called AnoGAN. A GAN was trained on normal medical images to learn the distribution of normal training data which later can be used to target anomalous samples. To do so, an optimization process on the random noise to find the closest generated image to the input test image was proposed. Albeit the model showed that a vanilla GAN could discriminate normal images from anomalous, it imposed considerable computation on the model leading to a very slow inference process. They defined a distance metric to measure how well a given test sample is generated as a way to discriminate anomalies.

A year later,~\citet{zenati2018efficient} presented an unsupervised model based on a bidirectional generative adversarial network (BiGAN) model~\citep{Donahue2017bigan, Dumoulin2017ali} with a similar scoring function as~\citet{schlegl2017unsupervised} to accelerate the inference procedure\footnote{For simplicity, we refer it as Efficient-GAN in the experiments and results section.}. 

Following the previous work, Zenati et al. proposed Adversarially Learned Anomaly Detection (ALAD) \citep{zenati2018adversarially}, a modification of their previous work, to detect anomalies. Their model contains three discriminators each receiving an input pair--one for handling the latent representations ($D_{zz}$), one for the input image $x$ ($D_{xx}$), and $D_{xz}$ which is similar to the discriminator used in BiGAN. For the inference, the $L_1$ reconstruction error in the feature space was used as the anomaly score:
\begin{equation}\label{ALAD_anomaly_score}
    A(x) = \norm{f_{xx}(x, x) - f_{xx}(x, G(E(x)))}_1
\end{equation}

\noindent where $f_{xx}$ is the activation of the layer before the logits in the $D_{xx}$ network, $E(x)$ is the representation obtained from the encoder $E$ for the given image $x$, and $G(E(x))$ is the output of the generator $G$ given $E(x)$.

\subsection{Our Anomaly Detection Model}\label{sec:model}
\begin{figure}%[b]%[htbp]
\centering
% \floatconts
\includegraphics[width=4.1in,height=1.2in]{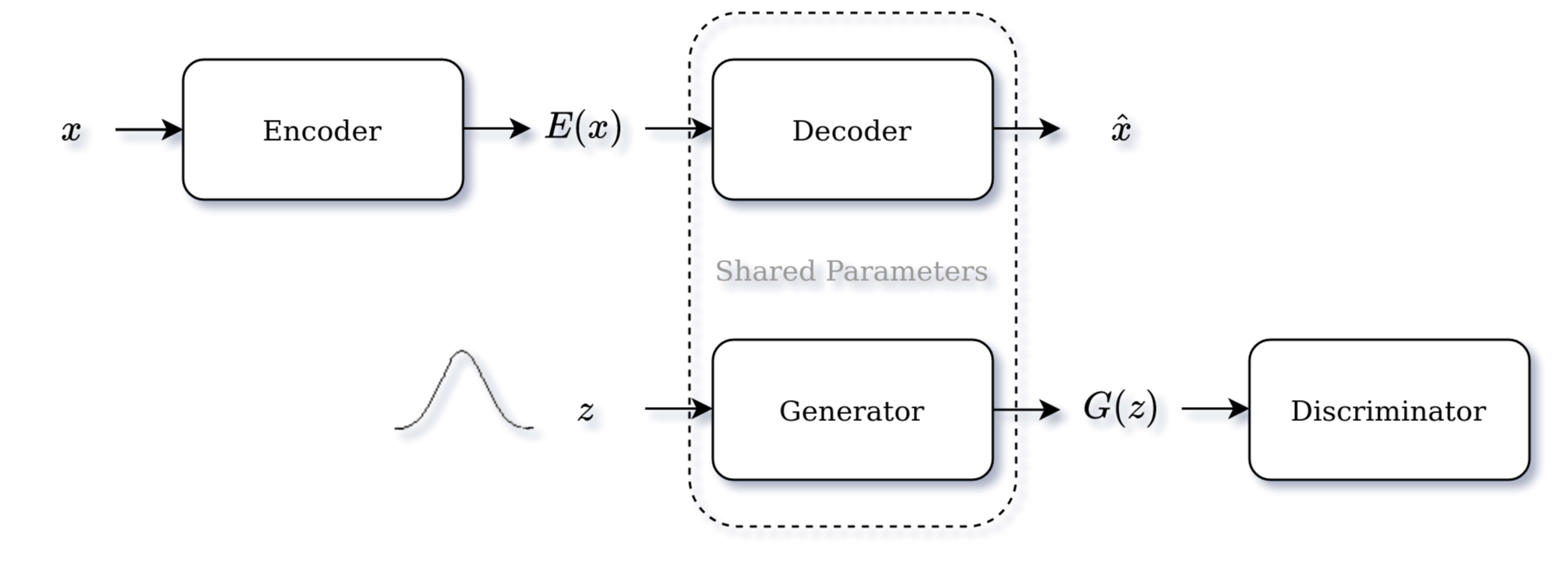}
\caption{\small{The GAN and autoencoder used in our model; encoder and discriminator have similar architecture except in their last layers, and the generator and the decoder share their weights.
}}
\label{fig:model}
\end{figure}

Similar to the previous AD models based on GANs, we suggest using adversarial training to identify anomalies.
%we propose an unsupervised model to detect anomalies using adversarial training.
We present a generative model, a combination of a GAN and an autoencoder (see Fig.~\ref{fig:model}). In this setting, we use parameter sharing (aka weight sharing) between GAN's generator and autoencoder's decoder to keep their distribution as close as possible. This will benefit the inference process by helping the encoder to generate representations within the distribution of the GAN. Our AD model trains on $D_{ind} = \{x_1, x_2, ..., x_k \sim P_{ind}\}$ where $P_{ind}$ defines normal (in-distribution) training samples. Therefore, the generated outputs of the GAN and the encoded representation of the encoder will be close to $P_{ind}$. During the inference, the model tests on $D_{mix} = \{x_1, x_2, ..., x_k \sim P_{ind}~or~P_{ood}\}$ where $P_{ood}$ defines anomalous (out-of-distribution) samples. Hence, the expected outputs of the GAN and the encoded representation of the autoencoder for an anomalous sample will be far from the actual test image and in another word close to $P_{ind}$. As a result, the dissimilarity between a given test sample and its corresponding generated output can be defined as our distance metric to target anomalous samples.

We train the GAN with relativistic standard GAN (RSGAN)~\citep{jolicoeur-martineau2018RSGAN} loss. Unlike the standard GAN (SGAN) objective function which measures the probability that the input data is real, Relativistic GAN measures the probability that the real data is more realistic than the generated data (or vice versa).
\begin{equation}
    \begin{split}
        L_D^{RSGAN} &= -\mathbb{E}_{(x_r, x_f) \sim (\mathbb{P}, \mathbb{Q})} [\log (sigmoid(C(x_r) - C(x_f)))]\\
        L_G^{RSGAN} &= -\mathbb{E}_{(x_r, x_f) \sim (\mathbb{P}, \mathbb{Q})} [\log (sigmoid(C(x_f) - C(x_r)))]
    \end{split}
\end{equation}
\noindent where $G$ and $D$ are the generator and discriminator of the GAN, $\mathbb{P}$ is the distribution of the real data, $\mathbb{Q}$ is the distribution of the fake data, $x_r$ and $x_f$ are real and fake data, and $C$ is the critic.

The autoencoder $AE$ was trained using the mean squared error (MSE) reconstruction loss function, $L_{AE} = \norm{x - G(E(x))}^2$, where $E(x)$ is the encoded representation of an input image $x$ produced by encoder $E$.

The anomaly score presented in this work modifies the previous scoring function presented in~\citep{schlegl2017unsupervised}.
\begin{equation}\label{eq:angoanscore}
    A(x) = \lambda L_D(x) + (1 - \lambda) L_R(x)
\end{equation}

As it is shown in Eq.~\ref{eq:angoanscore}, in~\citep{schlegl2017unsupervised}, the anomaly score of image $x$, $A(x)$, includes two terms--discrimination loss, $L_D(x)$, and residual loss, $L_R(x)$. These two terms compute the difference between the actual test image and its corresponding generated output from two different perspective. $L_D(x)$ relies on the intermediate representations ($f(\cdot)$) of them (Eq.~\ref{discrimimantion_loss}), while the $L_R(x)$ compute their visual dissimilarity (Eq.~\ref{residual_loss}).
\begin{equation}\label{discrimimantion_loss}
    L_D(x) = \sum |f_D(x) - f_D(G(E(x)))|
\end{equation}
\vspace{-15pt}
\begin{equation}\label{residual_loss}
    L_R(x)=\sum|x-G(E(x))|
\end{equation}

As stated earlier, we consider multiple representations of a single image to identify anomalies. Therefore, rather than discrimination loss and residual loss, we suggest using the encoded representation of the encoder as the latent loss, $L_L$ (shown in Eq.~\ref{eq:latentloss}). For a given image $x$, $L_L(x)$ compute how far the encoded representation of $x$, $E(x)$, is from the encoded representation of its generated output given $E(x)$.
\begin{equation}\label{eq:latentloss}
    L_L(x) = \sum |E(x) - E(G(E(x)))|
\end{equation}
By adding the latent loss to Eq.~\ref{eq:angoanscore}, we present a new anomaly score function, given in Eq.~\ref{eq:anomaly_score}. The effect of latent loss in our scoring function is controlled by the hyperparameter $\beta$.
\begin{equation}\label{eq:anomaly_score}
    A(x) = \lambda L_D(x) + (1 - \lambda) L_R(x) + \beta L_L(x)
\end{equation}
Given the anomaly score presented here, once the model learns the true distribution of normal (in-distribution) samples, $P_{ind}$, it will identify anomalous (out-of-distribution) samples, $P_{ood}$, by a higher anomaly score assigned to them as opposed to the score for the normal samples.

\section{Datasets}
To evaluate the performance of our model on AD tasks, in comparison with recent GAN-based models, we considered two types of datasets--natural images and medical images. MNIST~\citep{726791}, CIFAR10~\citep{Krizhevsky2009LearningML}, and SVHN~\citep{SVHN2011} as three benchmarks for natural images were chosen. For the medical dataset, we considered the Acute Lymphoblastic Leukemia (ALL) dataset~\citep{labati2011all} with only 260 images to evaluate our model's capability to perform under a limited data regime which is quite common in the medical domain.

Unlike the medical dataset, which provides normal and anomalous classes, each of the natural datasets has $10$ classes. Therefore each of those classes/labels separately can be defined as either normal or anomalous for our AD task. To this end, two new strategies to form the new datasets from the natural image datasets have been introduced here: 1) we define \textit{1 versus 9} where one out of $10$ classes is chosen to be anomalous while the rest form normal class, and 2) \textit{9 versus 1} where nine classes form the anomalous class and the remaining one form the normal class. These two strategies create $20$ different datasets for each of the natural datasets, with a total of $60$ datasets.

In the experiments on natural images, only normal images are considered for the training, while anomalous images and test data are used for the inference. In these experiments, a small proportion of samples is used as the validation sets. In order to evaluate the model on another domain with a fewer number of samples, the Acute Lymphoblastic Leukemia (ALL) dataset, with $260$ samples and an equal number of normal and anomalous samples for each class, is considered. From $D_{ind}$, $100$ samples are used for training, $20$ samples for the validation, and the remaining $140$ samples from $D_{mix}$, are considered for the inference.

\section{Experiments and Results}
The proposed model's performance was evaluated on natural (MNIST, CIFAR10, and SVHN) and medical (ALL) images. To be able to determine our model's benefits as well as its weaknesses, a comparison has been made on similar GAN-based AD models, Efficient-GAN~\citep{zenati2018efficient}, ALAD~\citep{zenati2018adversarially}, and AnoGAN~\citep{schlegl2017unsupervised}. Except for the AnoGAN, which suffers from a very long inference procedure (see Sec.~\ref{inference_time}), all the other models were evaluated on all four datasets.

% \begin{table}[t]
% \centering
% \caption{\small{Architecture and hyperparameters of our model on each dataset; the generator of the GAN and decoder of autoencoder use weight sharing. \textit{Conv} represents convolutional layers, \textit{Deconv} represents deconvolutional layers, and \textit{Linear} represents dense layers.}}\label{tab:models_arch}
% \vspace{1pt}
% \begin{adjustbox}{width=0.9\textwidth}
% \begin{tabular}{l|lll|lll|lll|lll}
% % \hline
% \multirow{2}{*}{Module} & \multicolumn{3}{c|}{ALL} & \multicolumn{3}{c|}{MNIST} & \multicolumn{3}{c}{CIFAR10} & \multicolumn{3}{c}{SVHN}\\ 
% & Conv & Deconv & Linear & Conv & Deconv & Linear & Conv & Deconv & Linear & Conv & Deconv & Linear \\ \hline
% Encoder  & 6 & - & 1 & 3 & - & 1 & 4 & - & - & 4 & - & - \\
% Generator/Decoder  & 1 & 8 & - & - & 3 & 2 & - & 4 & - & - & 4 & -\\
% Critic  & 6 & - & 1 & 3 & - & 2 & 3 & - & 1 & 3 & - & 1 \\
% \end{tabular}
% \end{adjustbox}
% \end{table}

\begin{table}[b]
    \centering
    \caption{\small{The architecture and hyperparameters of our model for the experiments on the MNIST, CIFAR10, SVHN and ALL datasets; the generator of the GAN and decoder of autoencoder use weight sharing. We used $i = 0, 0, 0, 1$, $ j = 3, 4, 4, 8$, $k = 2, 0, 0, 0$, $l = 3, 3, 3, 6$, $m = 2, 1, 1, 1$, $p = 3, 4, 4, 6$, and $q = 1, 0, 0, 1$ for MNIST, CIFAR10, SVHN and ALL dataset respectively.}}
    \vspace{3pt}
    \begin{adjustbox}{width=0.6\textwidth}
    \begin{tabular}{lcccc}
    \toprule
    & & \multicolumn{3}{c}{Our model architecture} \\
    \cmidrule(lr){3-5} 
    Module   & ~~ & $\#$Layers & Activation fn & Dropout \\
    \midrule
    $G(z)$   & ~~ &  \textbf{i} $\times$ $Conv2d$,  \textbf{j} $\times$ $Trans. Conv2d$, \textbf{k} $\times$ $Linear$ & $ReLU$ & $\times$ \\
    \midrule
    $D(x)$   & ~~ &  \textbf{l} $\times$ $Conv2d$, \textbf{m} $\times$ $Linear$ & $LeakyReLU$ & $0.2$ \\
    \midrule
    $E(x)$     & ~~ & \textbf{p} $\times$ $Conv2d$, \textbf{q} $\times$ $Linear$ & $LeakyReLU$ & $0.2$ \\
    \midrule
    \midrule
    % Latent dimension & ~~ & $100$ \\
    Learning rate & ~~ & $Lr_{GAN}$: $1 \times 10^{-4}$, $Lr_{AE}$: $1 \times 10^{-4}$ \\
    Optimizer & ~~ & Adam($\beta_1 = 0.5$, $\beta_2 = 0.999$)\\
    Batch Size & ~~ & $64$ (except ALL with $16$) \\
    % Weight initialization & ~~ & Spectral Normalization \\
    \bottomrule
    \end{tabular}
    \end{adjustbox}
    \label{tab:models_arch}
\end{table}

The detailed information of the choices of hyperparameters for our model on each of the experiments is indicated in Table~\ref{tab:models_arch}. For the medical domain, similar hyperparameters are used for all the GAN-based models compared in this study, while for the natural images, we used similar hyperparameters as presented in~\citep{zenati2018efficient, zenati2018adversarially}. In the case of the SVHN dataset, we compared our model with AnoGAN and ALAD following similar hyperparameters as~\citep{zenati2018adversarially}.
%However, they only reported their results for the \textit{9 versus 1} strategy.

For the experiment on the medical dataset, we trained each model for $1000$ epochs on the $D_{ind}$ with a learning rate of $1\text{e\ensuremath-}4$, batch size of $16$, latent size of $200$, and dropout ratio of $0.2$ for the encoder and discriminator. The models trained on natural images datasets for at most $85$ epochs, batch size of $64$, and learning rate of $1\text{e\ensuremath-}4$. The latent sizes of $100$ for MNIST and CIFAR10, and $200$ for SVHN were used, respectively. In all the experiments, models are optimized using the AdamW~\citep{loshchilov2018adamW} optimizer. During the inference, different values of $\beta$ were used for each dataset. These values were determined experimentally and defined the contribution of the latent loss in the new anomaly score. Specifically, $\beta = 1$ for CIFAR10, SVHN, and ALL datasets and $\beta = 0.5$ for MNIST dataset were used. $\lambda=0.8$ was chosen experimentally for all the experiments.

\subsection{Experimental Setup}
% \subsubsection{Residual Loss versus Discrimination Loss}\label{lambda-experiments}
\subsubsection{The impact of $\lambda$}\label{lambda-experiments}
\begin{figure}[t] %[hbp] %[bp]
  \centering
  \subfloat{\includegraphics[width=37mm,height=32mm]{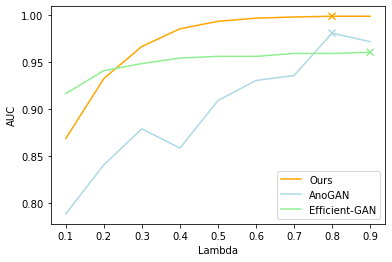}}
  \subfloat{\includegraphics[width=37mm,height=32mm]{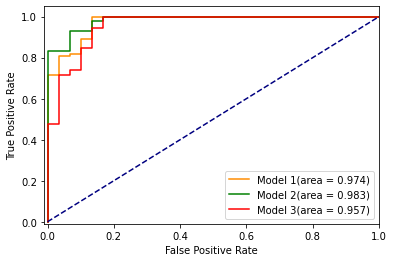}}
  \subfloat{\includegraphics[width=37mm,height=32mm]{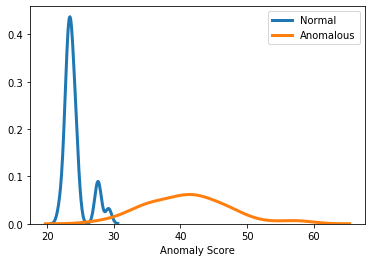}}
  \caption{\small{In the left, the performance of Efficient-GAN, AnoGAN, and Ours for different contributions of discrimination and residual losses under coefficient $\lambda$ are depicted. In the middle and right, the performance of our model on three runs with their ROC curves along with the anomaly score distribution of our best models out of three runs are shown.
  }} \label{fig:lambda_and_ROC_compariosn}
\end{figure}

One of the key factors in the performance of the recent GAN-based models is the effectiveness of their scoring function. In~\citep{schlegl2017unsupervised, zenati2018efficient} as well as our model, different contributions of the learned features of the critic (discrimination loss) and the visual dissimilarity of the generated samples and actual test samples (residual loss) in the final scoring can have a huge impact on the performance of each of these models. In a small experiment on the ALL dataset, the effect of different values of $\lambda$ in the range of $[0, 1]$ on the performance of Efficient-GAN, AnoGAN, and our model was investigated. These models were compared based on their  area under the ROC (receiver operating characteristic) curve (AUC).
 For the experiment on our modified scoring function, we used a fixed value of $1$ for $\beta$. It can be observed from Fig.~\ref{fig:lambda_and_ROC_compariosn} that all these models perform better with larger $\lambda$ indicating a higher contribution of discrimination loss.

Since the residual loss is more sensitive to the artifacts in the generated output, comparing the internal representation of a given image might ignore those visual differences and focus on more abstract features. More detail on the effectiveness and challenges of using these two losses is explained in the analysis section (Sec~\ref{sec_analysis}).

\subsubsection{Stabilizing Training}
One of the biggest challenges in training GANs are their instability. Slight changes in model's hyperparameters, running on different machines, and even random initialization can affect their performance more than any other deep models~\citep{lucic2018gans}. Therefore to reduce the instability of our model during training, spectral normalization~\citep{miyato2018spectral} was used for the critic. To compare the model's performance independent of its random initialization, the model was trained three times with different random initializations. All of the results reported in this study were computed as an average on the three runs from these different random initializations.

\subsection{Experimental Results}

\subsubsection{Medical Imaging Dataset}\label{sec:ALL_result}
The detailed performance of all four GAN-based models on our medical imaging dataset is summarized in Table~\ref{Table:models_result_our_way_ALL}. As illustrated in the Table, Ours showed a high capability to detect anomalies from various performance metrics. Ours outperformed the existing approaches on AUC with a large margin (increased by $\%10$). In terms of specificity, the best performance is acquired by Efficient-GAN with Ours as the second best.

\begin{table}[t]
\centering
\caption{\small{The AUC ($\%$) comparison on the ALL dataset for AnoGAN, Efficient-GAN, and our model with $0.8$, $0.9$, and $0.8$ for coefficient $\lambda$ for each method, respectively. In this and the following tables, the results obtained from our implementation are represented by the $^\intercal$ sign. ($\pm$ std.~dev.)
\vspace{4pt}}}\label{Table:models_result_our_way_ALL}
\begin{adjustbox}{width=0.87\textwidth}
\begin{tabular}{l|c|c|c|c|c}
Model & Sensitivity & Specificity & f1-measure & Accuracy & AUC \\ \hline
AnoGAN$^\intercal$  & 73.08~$\pm$~0.254 & 74.44~$\pm$~0.164 & 79.19~$\pm$~0.203 & 73.34~$\pm$~0.236 & 75.71~$\pm$~0.241 \\
Efficient-GAN$^\intercal$ & 71.54~$\pm$~0.229 & \textbf{98.89}~$\pm$~0.016 & 81.07~$\pm$~0.165 & 76.67~$\pm$~0.183 & 87.23~$\pm$~0.137\\
ALAD$^\intercal$ & 94.61~$\pm$~0.016 & 75.0~$\pm$~0.057 & 88.52~$\pm$~0.016 & 86.09~$\pm$~0.022 & 79.88~$\pm$~0.048 \\
Ours & \textbf{98.72}~$\pm$~0.004 & 84.44~$\pm$~0.016 & \textbf{97.73}~$\pm$~0.001 & \textbf{96.04}~$\pm$~0.003 & \textbf{97.31}~$\pm$~0.009
% Ours & \textbf{0.9615}~$\pm$~0.029 & 0.9222~$\pm$~0.041 & \textbf{0.9713}~$\pm$~0.011 & \textbf{0.9542}~$\pm$~0.016 & \textbf{0.9893}~$\pm$~0.010
\end{tabular}
\end{adjustbox}
\end{table}

The observation on the range of standard deviation from multiple runs showed that AnoGAN had the least stability. In comparison, the highest stability is achieved by Ours, which can be inferred from both ROC curves of Ours on three runs (Fig.~\ref{fig:lambda_and_ROC_compariosn}, middle plot) and the results from Table~\ref{Table:models_result_our_way_ALL}. We also showed that our model could effectively discriminate normal and anomalous samples even on a very small dataset (Fig.~\ref{fig:lambda_and_ROC_compariosn}, third plot from the left).

\begin{figure}[b] %[htb]
  \centering
  \subfloat{\includegraphics[width=50mm,height=34mm]{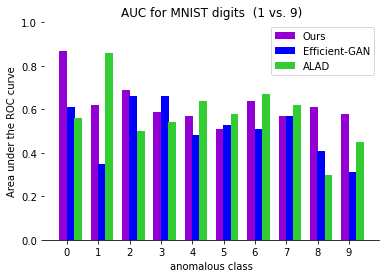}}
  \subfloat{\includegraphics[width=50mm,height=34mm]{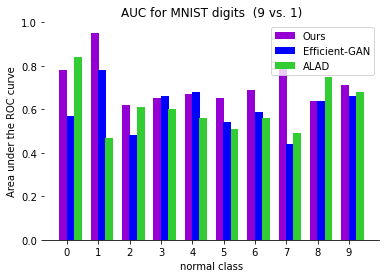}}
  \vskip\baselineskip
  \vspace{-0.6cm}
  \subfloat{\includegraphics[width=50mm,height=40mm]{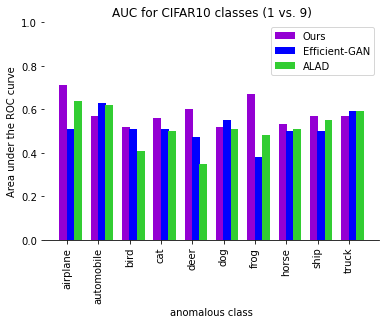}}
  \subfloat{\includegraphics[width=50mm,height=40mm]{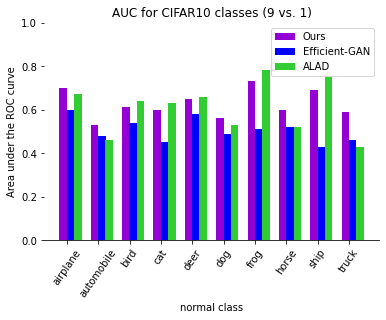}}
  \caption{\small{Individual performance of each label on MNIST and CIFAR10.}}
  \label{fig_MNIST_CIFAR10_details}
\end{figure}

% \begin{figure}[b] %[htb]
%   \centering
%   \subfloat{\includegraphics[width=50mm,height=34mm]{images/MNIST_cmp_1vs9.png}}
%   \subfloat{\includegraphics[width=50mm,height=34mm]{images/MNIST_cmp_9vs1.png}}
%   \vskip\baselineskip
%   \vspace{-0.6cm}
%   \subfloat{\includegraphics[width=50mm,height=40mm]{images/CIFAR10_cmp_1vs9.png}}
%   \subfloat{\includegraphics[width=50mm,height=40mm]{images/CIFAR10_cmp_9vs1.png}}
%   \caption{\small{Individual performance of each label on MNIST and CIFAR10.}}
%   \label{fig_MNIST_CIFAR10_details}
% \end{figure}

\subsubsection{Natural Images}
For our experiments on natural images, we considered the aforementioned \textit{9 versus 1} and \textit{1 versus 9} strategies and compared the performance of our model with Efficient-GAN~\citep{zenati2018efficient} and ALAD~\citep{zenati2018adversarially}. Table~\ref{Table:models_result_our_way} summarizes the AUC of each model within each of these strategies, which are averaged over three runs on all the classes of MNIST and CIFAR10. As the results reveal, our model outperformed the other two GAN-based models on all the experiments by a large margin. The detailed results of all three compared models on each of the classes of MNIST and CIFAR10 are also depicted in Fig.~\ref{fig_MNIST_CIFAR10_details}.

\begin{table}[!htb]
    \caption{\small{The AUC ($\%$) comparison on MNIST and CIFAR10 datasets with \textit{1 versus 9} and \textit{9 versus 1} strategies. Results from the original papers are indicated by $^\star$ symbol. ($\pm$ std.~dev.)\vspace{4pt}
    }}
    \centering
    \begin{adjustbox}{width=0.7\textwidth}
      \begin{tabular}{r@{\extracolsep{3pt}}*{6}{c}}
        \toprule
        & \multicolumn{2}{c}{\textit{1 versus 9}} &~~& \multicolumn{2}{c}{\textit{9 versus 1}} \\
        \cmidrule(lr){2-3} \cmidrule(lr){5-6}
        Model & MNIST & CIFAR10 && MNIST & CIFAR10 \\
        \midrule
Efficient-GAN$^\intercal$ & 50.9 $\pm$ 0.116 & 51.5 $\pm$ 0.064 && 60.4 $\pm$ 0.096 & 50.6 $\pm$ 0.053\\
ALAD$^\intercal$ & 57.2 $\pm$ 0.140 & 51.6 $\pm$ 0.086 && 60.7 $\pm$ 0.112 & ~60.7 $\pm$ 0.120$^\star$\\
        \midrule
Ours & \textbf{62.5} $\pm$ 0.093 &\textbf{58.2} $\pm$ 0.060 && \textbf{71.6} $\pm$ 0.096 & \textbf{62.6} $\pm$ 0.061 \\
        \bottomrule
      \end{tabular}
    \end{adjustbox}
    \label{Table:models_result_our_way}
\end{table}

\begin{table}[t] %[!htb]
\centering
\caption{\small{The AUC ($\%$) of AnoGAN, ALAD and our model on SVHN dataset with \textit{1 versus 9} and \textit{9 versus 1} strategies.}}\label{tab:SVHN_results}
\begin{adjustbox}{width=0.45\textwidth}
\begin{tabular}{lccc}
\toprule
\small{Model} & \textit{1 versus 9} & ~~ & \textit{9 versus 1} \\
\midrule
\small{AnoGAN} & 46.6 $\pm$~1.3 & & ~54.1~~$\pm$~0.019 \\
\small{ALAD} & ~~51.6 $\pm$~0.09 & & ~57.5$^\star$$\pm$~0.027 \\
\midrule
% Ours (SGAN) & 0.568$\pm$~0.007 &\textbf{0.581}~$\pm$~0.014 \\
\small{Ours} & ~~~~\textbf{56.8}~$\pm$~0.007 & & ~\textbf{58.1}~~$\pm$~0.014 \\
\bottomrule
\end{tabular}
\end{adjustbox}
\end{table}

On the SVHN dataset, we evaluated the performance of our model on both \textit{9 versus 1} and \textit{1 versus 9} strategies. As shown in Table~\ref{tab:SVHN_results}, our model outperformed its two other rivals on \textit{1 versus 9} strategy with at least $\%5$ improvement on AUC. The results however indicate a slight improvement (less than $\%1$) in the performance of our model on \textit{9 versus 1} in comparison with ALAD.

% On the SVHN dataset, we evaluated the performance of our model on both \textit{9 versus 1} and \textit{1 versus 9} strategies. However, as shown in Table~\ref{tab:SVHN_results}, the comparison with AnoGAN and ALAD was only done on \textit{9 versus 1} strategy which indicates a similar performance of our model compared with ALAD with larger improvement over AnoGAN.

\subsubsection{Output Analysis}\label{sec_analysis}
\begin{figure}[b] %[!htb]
  \centering
  \subfloat[Outputs of \textit{9 versus 1} strategy on SVHN and CIFAR10 datasets]{\includegraphics[width=45mm,height=20mm]{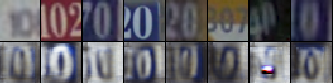}
  \includegraphics[width=45mm,height=20mm]{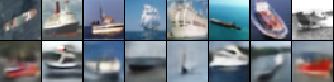}}
  \vskip\baselineskip
  \vspace{-0.7cm}
  \subfloat[Outputs of \textit{1 versus 9} strategy on MNIST and CIFAR10 datasets]{\includegraphics[width=45mm,height=20mm]{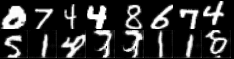}
  \includegraphics[width=45mm,height=20mm]{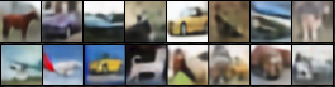}}
  \caption{\small{The generated outputs of our model on SVHN, CIFAR10, and MNIST datasets using \textit{9 versus 1} and \textit{1 versus 9} strategies. The top rows of each sub-figure (a) and (b) show the training images, and the second rows are the generated images by the GAN.}}
  \label{fig_outputs}
\end{figure}

A thorough analysis of the generated outputs of our model on different datasets revealed that in the case of \textit{9 versus 1} when only one of the labels form the normal class, the model is better able to capture the distribution of the normal data which is reasonable considering the model is learning an easier pattern. Even though this is the case for almost all the datasets (Fig.~\ref{fig_outputs} (a) right; \textit{9 versus 1} on CIFAR10), the model has difficulty when training on the SVHN dataset even when it should learn the distribution of just a single label representing the normal class. This is mostly due to the nature of this dataset where the classes are not completely separated, i.e., the samples of the class zero can contain other digits in their image (Fig.~\ref{fig_outputs} (a) left) which makes it hard for the model to learn the true distribution of the digit zero. This phenomenon can affect the performance, especially during inference time where the visual dissimilarity of the generated image and the actual test image can have a direct impact on identifying anomalous samples. The model can easily fail and even if the model is able to generate the test digit, there can be often visual artifacts causing high residual loss.

We also observed that, in the cases where digits with similar patterns are considered as the normal class (with \textit{9 versus 1} strategy), the model may fail to identify the anomalous image when the corresponding test image has a similar pattern. For instance, when considering digit $3$ as the normal sample, the model can fail when the actual test image is digit $8$, hence, receiving lower residual and discrimination losses and therefore will be recognized as a normal sample.

Considering \textit{1 versus 9} strategy where $9$ classes form the normal training data, mode collapse was the major issue in training the model for our anomaly detection purpose. As an example, in Fig.~\ref{fig_outputs} (b), the model is more focused on learning the distribution of cars and planes in CIFAR10 dataset and digit seven and digit one while training on MNIST dataset and ignores the other classes. As the result, it may fail to learn the whole distribution while focusing on only a subset of the training distribution, therefore leading to high anomaly scores for the samples actually coming from the normal training distribution.

\subsubsection{Inference Time Comparison}\label{inference_time}
One of the major challenges in training a vanilla GAN for anomaly detection is its long inference process which negatively affects required time and computational resources for performance. Therefore, we modified the GAN by adding an autoencoder to help the model improve the existing results while reducing the inference time. A comparison on all the GAN-based models studied in this work on the ALL dataset with $160$ test images is shown in Table~\ref{tab:models_inference_time}. As observed from the Table, Ours slightly improved the inference time compared to \citep{zenati2018efficient} and \citep{zenati2018adversarially}, while the improvement is more notable compared to \citep{schlegl2017unsupervised}. Python 3.7 with the PyTorch~\citep{NEURIPS2019_9015} library on a GeForce GTX 1080 Ti GPU was used for these experiments. We considered $500$ iterations for AnoGAN to optimize the random noise $z$ for each given test image.

\begin{table}[t] %[!htb]
\centering
\caption{\small{Inference time comparison on the ALL dataset on images of shape (3, 220, 220) with a (200, 1) vector of noise randomized from a Gaussian distribution.
\vspace{4pt}}}\label{tab:models_inference_time}
\begin{adjustbox}{width=0.7\textwidth}
\begin{tabular}{l|c|c|c|c}
\multicolumn{1}{c|}{\multirow{2}{*}{Models}} & \multicolumn{3}{c|}{$\#$ of parameters in each module} & \multicolumn{1}{c}{\multirow{2}{*}{Inference time (ms)}} \\
% \cline{2-4}
\multicolumn{1}{c|}{} & \multicolumn{1}{c|}{Encoder} & \multicolumn{1}{c|}{Decoder/Generator} & \multicolumn{1}{c|}{Critic} & \multicolumn{1}{c}{} \\ \cline{1-1} \cline{2-4} \cline{5-5} 
AnoGAN$^\intercal$  & - & 2,450,307 & 5,159,170 & 13110.47 \\
Efficient-GAN$^\intercal$  & 5,874,352 & 1,906,240 & 7,024,929 & 3.33 \\
ALAD$^\intercal$  & 5,771,752 & 1,906,240 & 7,814,915 & 3.85 \\
Ours  & 8,716,888 & 2,450,307 & 5,159,170 & 2.90
\end{tabular}
\end{adjustbox}
\end{table}

\subsection{Ablation Study}

\subsubsection{Latent Loss Impact}
The new anomaly score presented in this work is a modification of an existing scoring function~\citep{schlegl2017unsupervised, zenati2018efficient} where we try to leverage the learned features of the autoencoder. Therefore to show the effectiveness of the new anomaly score, a comparison on the natural images using the new and original anomaly score was conducted. The results on Table~\ref{tab:latent_loss_compare} demonstrate the benefit of the added latent loss in the new anomaly score.

\begin{table}[!htp] %[t] 
\centering
\caption{\small{The effect of latent loss in the new anomaly score. The comparison were done on natural images. In the experiments using latent loss, $0.5$ and $1$ were used as $\beta$ for MNIST and CIFAR10 respectively. ($\pm$ std.~dev.)\vspace{4pt}}} \label{tab:latent_loss_compare}
\begin{adjustbox}{width=0.8\textwidth}
\begin{tabular}{l|ccc|cc}
& \multicolumn{2}{c}{\textit{1 versus 9}} &~~& \multicolumn{2}{c}{\textit{9 versus 1}} \\
Model & MNIST & CIFAR10 && MNIST & CIFAR10\\ \hline
without latent loss ($\beta = 0$) & 54.7 $\pm$ 0.099 & 50.2 $\pm$ 0.084 && 67.0 $\pm$ 0.114 & 56.9 $\pm$ 0.107 \\
with latent loss ($\beta \neq 0$) & \textbf{62.5} $\pm$ 0.093 &\textbf{58.2} $\pm$ 0.060 && \textbf{71.6} $\pm$ 0.096 & \textbf{62.6} $\pm$ 0.061 \\
\end{tabular}
\end{adjustbox}
\end{table}

\subsubsection{GAN Objective}
To have a better understanding of the effectiveness of Relativistic GAN loss for our model, two different losses for GAN have been considered. Precisely, RSGAN and SGAN objective functions were compared on the natural datasets experimented on in this study. In all of the experiments, using RSGAN increased the performance of our model (see Table~\ref{tab:GAN_objective}).
\begin{table}[!htp]
\centering
\caption{\small{The effect of using different GAN objective functions on the performance of our model. ($\pm$ std.~dev.)\vspace{4pt}}} \label{tab:GAN_objective}
\begin{adjustbox}{width=0.8\textwidth}
\begin{tabular}{l|ccc|cc}
& \multicolumn{2}{c}{\textit{1 versus 9}} &~~& \multicolumn{2}{c}{\textit{9 versus 1}} \\
GAN objective fn. & MNIST & CIFAR10 && MNIST & CIFAR10\\ \hline
Standard GAN (SGAN) & 55.7 $\pm$ 0.075 & 55.5 $\pm$ 0.082 && 69.3 $\pm$ 0.129 & 61.1 $\pm$ 0.088 \\
Relativistic GAN (SRGAN) & \textbf{62.5} $\pm$ 0.093 &\textbf{58.2} $\pm$ 0.060 && \textbf{71.6} $\pm$ 0.096 & \textbf{62.6} $\pm$ 0.061 \\
\end{tabular}
\end{adjustbox}
\end{table}

\section{Conclusion and Future Work}
In this work, we suggested using a simple and effective generative model to identify anomalies. The model contains a GAN and an autoencoder, which train simultaneously on the normal training data. To detect anomalies during inference time, we introduced a new anomaly score function comprising multiple representations obtained from the autoencoder and the GAN. We further evaluated our model on MNIST, CIFAR10, SVHN, and a public Acute Lymphoblastic Leukemia (ALL) datasets. Our model proved its performance in all of the experiments with a large improvement over the existing GAN-based models with lower inference time. We also showed that our model could perform quite well even on small-sized datasets. Despite the effectiveness of our model on identifying anomalies, mitigating the challenges in training GANs and learning more complicated distribution seem to be necessary. To this end, in our future work, we tend to study the effect of using contrastive learning in training GANs to learn more discriminative representations of the images while investigating different scoring functions to fill this gap.

\bibliographystyle{unsrtnat}
\bibliography{references}  %%% Uncomment this line and comment out the ``thebibliography'' section below to use the external .bib file (using bibtex) .





% \end{thebibliography}

\end{document}